\documentclass{article}

\usepackage{microtype}
\usepackage{graphicx}
\usepackage{subfigure}
\usepackage{booktabs} % for professional tables

\usepackage{hyperref}

\usepackage[accepted]{icml2023}

\usepackage{amsmath}
\usepackage{amssymb}
\usepackage{mathtools}

\DeclarePairedDelimiterX{\infdivx}[2]{(}{)}{%
  #1\;\delimsize\|\;#2%
}

\usepackage{amsthm}
\usepackage{bm}

\usepackage[capitalize,noabbrev]{cleveref}

\theoremstyle{plain}

\theoremstyle{definition}

\theoremstyle{remark}

\usepackage[textsize=tiny]{todonotes}

\icmltitlerunning{Online Learning for the Random Feature Model in the Student-Teacher Framework }

\begin{document}

\twocolumn[
\icmltitle{Online Learning for the Random Feature Model in the Student-Teacher Framework }



\icmlsetsymbol{equal}{*}

\begin{icmlauthorlist}
\icmlauthor{Roman Worschech}{yyy,comp}
\icmlauthor{Bernd Rosenow}{comp}
%\icmlauthor{Firstname3 Lastname3}{comp}
%\icmlauthor{Firstname4 Lastname4}{sch}
%\icmlauthor{Firstname5 Lastname5}{yyy}
%\icmlauthor{Firstname6 Lastname6}{sch,yyy,comp}
%\icmlauthor{Firstname7 Lastname7}{comp}
%\icmlauthor{Firstname8 Lastname8}{sch}
%\icmlauthor{Firstname8 Lastname8}{yyy,comp}
\end{icmlauthorlist}

\icmlaffiliation{yyy}{Max Planck Institute for Mathematics in the Sciences, D-04103 Leipzig, Germany}
\icmlaffiliation{comp}{Institut f{\"u}r Theoretische Physik, Universit{\"a}t Leipzig, Br{\"u}derstrasse 16, 04103 Leipzig, Germany}
%\icmlaffiliation{sch}{School of ZZZ, Institute of WWW, Location, Country}

\icmlcorrespondingauthor{Roman Worschech}{roman.worschech@mis.mpg.de}
%\icmlcorrespondingauthor{Firstname2 Lastname2}{first2.last2@www.uk}

\icmlkeywords{Machine Learning, ICML}

\vskip 0.3in
]



\printAffiliationsAndNotice{}  % leave blank if no need to mention equal contribution

\begin{abstract}
Deep neural networks are widely used prediction algorithms whose performance often improves as the number of weights increases, leading to over-parametrization. We consider a two-layered neural network whose first layer is frozen while the last layer is trainable, known as the random feature model. We study over-parametrization in the context of a student-teacher framework by deriving a set of differential equations for the learning dynamics. For any finite ratio of hidden layer size and input dimension, the student cannot generalize perfectly, and we compute the non-zero asymptotic generalization error. Only when the student's hidden layer size is exponentially larger than the input dimension, an approach to perfect generalization is possible.
\end{abstract}

\section{Introduction}
Deep neural networks (DNNs) have a wide range of applications. Besides the most well-known areas, such as image classification and speech recognition, there are numerous applications within physics \cite{lecun2015deep,Goodfellow2016,Kri+12,Sil+17,Carleo+2019}. Often, DNN performance improves as the number of weights increases, generating networks that have far more free parameters available than training data, a regime which reveals exciting properties \cite{Zhang+2017, novak2018sensitivity, canziani2017, novak2019bayesian, neyshabur2018the, Bartlett98}. 
When initializing the weights randomly and independently with zero mean and a variance inversely proportional to the network width, the network's output in the infinite width limit is described by a Gaussian process with an analytically accessible covariance matrix \cite{Neal1996priors, lee2018deep, novak2019bayesian, matthews2018gaussian, Williams96}. However, the evolution of network weights during training and the corresponding improvement of prediction performance are not fully understood yet. \\

In the ultra-wide limit for which the neural network size tends to infinity, recent studies have revealed that over-parametrized neural networks trained by stochastic gradient descent can be described by two different limits, depending on the scaling of the output layer \cite{Geiger20}. In order to realize the so-called neural tangent kernel (NTK) limit, the network output is scaled by the square root of the network width. In this limit, the stochastic gradient descent changes the weights of the network very slowly, resulting in weigth dynamics equivalent to that of a linearized network, and the evolution of network predictions during the learning process is described by the so-called neural tangent kernel  \cite{Jacot+18}. This limit is also known as the lazy training regime, for which the network weights stay in the vicinity of their initial values during the learning process, and for which the NTK does not change during training in the ultra-wide limit \cite{Chizat2019on}.  \\

Here, we study a student-teacher secanario in the NTK limit, which allows us to analyze the generalization performance and convergence properties of wide over-parametrized neural networks \cite{Arora2}. However, it is not entirely clear whether the NTK framework can fully explain the performance of DNNs. Using certain data sets, the autors of \cite{Arora+19} showed that the NTK limit achieves good generalization performance, and in \cite{Lee2019wide} it was found found that the prediction of the NTK limit is in correspondence with certain over-parametrized and in-practice used neural networks. In contrast, \cite{Chizat2019on,Ghorbani2019limitation,Yehudai2019on} highlighted that the NTK limit cannot explain the generalization performance of over-parametrized neural networks since a performance gap was revealed. This performance gap was recently confirmed by \cite{Samarin22} empirically for finite and practically relevant networks. In this work, we want to further quantify this performance gap analytically and analyze under which circumstances a DNN in the NTK limit can be expected to perform well.\\
Throughout this paper, we consider two-layer neural networks in the lazy training regime, 
%and use a Taylor expansion with regards to changes in the network weights, 
resulting in a linear learning dynamics described by the NTK \cite{Rahimi2008}. 
In order to study the behavior of neural networks in the NTK limit qualitatively, we freeze the first layer weights mimicking their laziness and train only the hidden-to-output neurons. Such a network architecture is known as the random feature model (RFM) \cite{Rahimi2007}, where the feature map is equivalent to the activations of the hidden neurons. The RFM corresponds to the part of a fully linearized two-layer neural network where one linearizes the network's output for hidden-to-output weights only and neglects the linearization with respect to input-to-hidden weights. However, on a qualitative level the results for a RFM and the full NTK limit may be similar since freezing the input-to-hidden neurons may not affect the predictors class \footnote[1]{Note that non-trivial quantitative differences exists between RFM and the fully NTK limit when the objective function is a low-order polynomial as the fully NTK linearization has more regression parameters than the RFM \cite{Ghorbani2019limitation}.}. \\
In this paper, we use techniques from statistical mechanics to analyze the learning dynamics of highly over-aprametrized random feature models with a large hidden and input layer. The network is trained by one-pass stochastic gradient descent, defining the system's dynamics. In order to express over-parametrization, we use the student-teacher framework for which the student is trained by the outputs of a teacher \cite{Gardner1989}. In this setup, the number of student hidden nodes $K$ is larger or equal to the number of teacher hidden nodes $M$. In such a case, the student is potentially more powerful than the teacher and can express more complex functions. Moreover, we are interested in the influence of the ratio of hidden-layer width $K$ to input-layer size $N$ on the student performance. Since over-parametrized networks have many degrees of freedom, statistical mechanics forms an important branch of theory deriving macroscopic properties from the interaction of network weights \cite{hertz1991introduction,Wat_92+,Saad98,Engel2001,Bahri+20}. It was successfully used to study the perceptron \cite{Gardner88,gyorgyi1990neural,Seung+92}, two-layer neural networks \cite{Opper94,riegler1995line,goldt2019dynamics,Schwarze+93,biehl1995learning,saad1995line,saad1995exact,Biehltransient,Biehl+98,straat2019line,Richert2022}, and more general (deep) architectures \cite{Naveh+20,Li2021statistical,Cohen+21}. Other interesting investigations deal with the influence of the input structure on the learning process \cite{Goldt+20,Yoshida_2020}.

\section{Related work}
Based on the bias-variance tradeoff, traditional learning theories suggest that over-parametrized neural networks should generalize poorly as they have more regression parameters than avaiable input examples \cite{Geman1992neural}. However, recent studies have shown that this tradeoff must be extend for ultra-wide neural networks leading to a double descent curve \cite{Belkin+19}. For the RFM, different analytical studies derived the test error's double descent behavior in the student-teacher framework indicating the accessibility of interesting deep learning phenomena by investigating this model \cite{advani2020high,D'Ascoli2020, Mei2022the, Hastie22}. \\

Related work has shown that the RFM cannot predict the output of a ReLU perceptron, except the hidden layer size $K$ increases exponentially fast with the input dimension $N$ \cite{Yehudai2019on}. We verify this result for an error activation function by actually computing the asymptotic generalization error. Further related studies considered the random feature model for quadratic activation and output functions in the high-dimensional regime $K,N \rightarrow \infty$, $K/N \rightarrow \tau \in \left(0,\infty\right)$ trained by one-pass stochastic gradient descent \cite{Ghorbani2019limitation}. These works demonstrated the existence of a gap between the performance of the RFM compared to fully-trained over-parametrized neural networks. \cite{Mei2022Generalization} provide a general expresssion for the generalization error of random feature models, relating it to the projection of the target function on the span of kernel eigenfunctions which are not learned in the regime of $K/N $-finite. Here, we provide an explicit expression for the $K/N $ dependence of the generalization error for the case of the error and ReLU activation function. Furthermore, the network performance is highly controlled by the initialization of the weights and becomes optimal only in the case for which the random features are perfectly aligned with those of the teacher. For $K \propto N$ \cite{ Ghorbani2021linearized} have shown that the student is just able to learn linear teachers. This regime lies in our focus and justifies the linearization of the random feature correlation matrices introduced in Chapter 5. We further extend investigations by introducing a set of ordinary differential equations for the learning dynamics of a student with infinitely large input and hidden layer size trained by a teacher with a ReLU or error activation function. We find that the generalization error converges to a finite plateau whenever the ratio of the input-to-hidden layer size is finite and compute the plateau heights for different activation functions rather than providing error bounds.

\section{Setup}
For our setup, the student and teacher receive input samples $\bm{\xi}^\mu\in\mathbb{R}^N$ presented once and in a sequential order where each component is generated by the normal distribution $\xi_i^\mu\in\mathcal{N}\left(0,1\right)$ with $\mu \in \{1,...,p\}$. The student has $K$ hidden neurons and the connection between the input layer with the $i$-th hidden node is expressed by the student vectors $\bm{J}_i\in\mathbb{R}^N$. The outputs of the hidden nodes are given by a continuous activation function $g$. The overall output of the network is a linear combination of the outputs of the hidden units 
\begin{align}
\sigma\left(\bm{J},\bm{\xi}\right)=\sum^K_i c_i g\left(x_i\right),
\end{align}
with $c_i$ are the hidden-to-output weights and the pre-activations of the student are defined by $x_i=\frac{\bm{J}_i\cdot\bm{\xi}}{\sqrt{N}}$. Note that the rescaling factor $\frac{1}{\sqrt{N}}$ guarantees pre-activations of $O\left(1\right)$. The teacher has $M$ hidden neurons characterized by the teacher vectors $\bm{B}_n\in\mathbb{R}^N$ and provides the output $\zeta(\bm{B},\bm{\xi})=\frac{1}{\sqrt{M}}\sum_{n=1}^{M} g\left(y_n\right)$ with $y_n=\frac{\bm{B}_n\cdot\bm{\xi}}{\sqrt{N}}$ as the pre-activations of the teacher. However, it turns out that increasing the teacher size $M$ does not influence the learning process qualitatively and we therefore consider a teacher perceptron with $M=1$.\\
\begin{figure}
\centering
\includegraphics[width=1\linewidth]{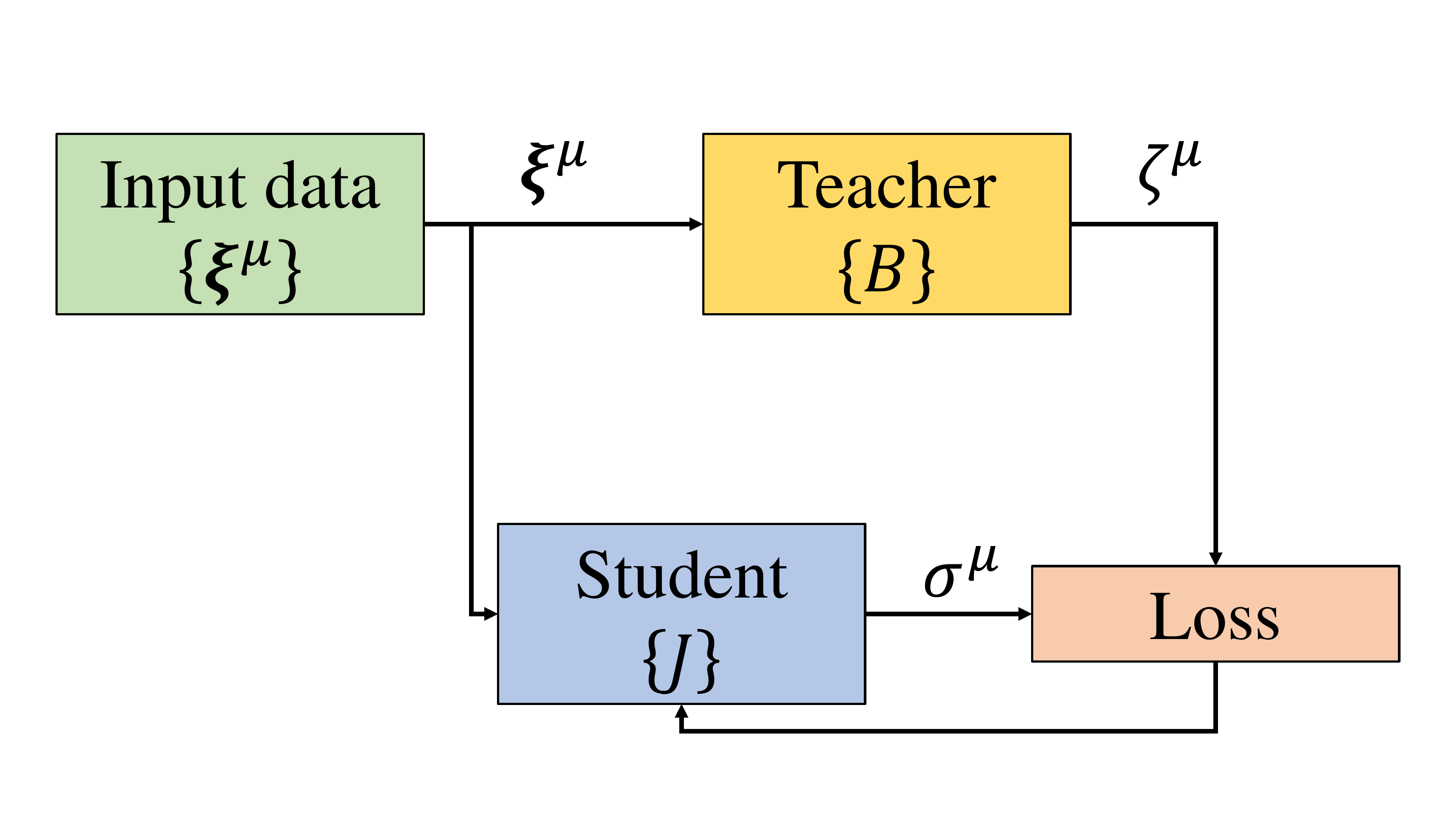}
\caption[Learning Problem]{Schematic illustration of the learning problem. An environment generates input data $\{\bm{\xi}^\mu\}$ according to a probability distribution. The student and teacher receive this data. Both release their outputs according to their parameter sets $\{J\}$ and $\{B\}$, respectively. The loss quantifies the dissimilarity between the student output $\sigma^\mu$ and teacher output $\zeta^\mu$. It sends feedback to the student for the adaptation of the parameters $\{J\}$. The overall goal is to minimize the loss during the learning process.}
\label{fig:LM}
\end{figure}
We choose the student and teacher vectors independently from the uniform distribution over the $N$ sphere and initialize the output weights of the student by the normal distribution with $c_i\in\mathcal{N}\left(0,\frac{1}{K}\right)$. In practice, we initialize the components of the student and teacher vectors by the normal distribution $J_i ,B_n\in\mathcal{N}\left(0,1\right)$ and rescale them afterwards by their norm. This leads to a uniform distribution of the student and teacher vectors over the $N$ sphere with radius $\sqrt{N}$ making the student vectors to random features. As we are interested in cases for which $K>N$, the student vectors are expected to have important correlations among each other. In order to express these correlation and those of the student and teacher vectors we introduce the variables $R_{in}=\frac{\boldsymbol{J}_i\cdot\boldsymbol{B}_n}{N},Q_{ij}=\frac{\boldsymbol{J}_i\cdot\boldsymbol{J}_j}{N}$ and $T_{nm}=\frac{\boldsymbol{B}_n\cdot\boldsymbol{B}_m}{N}=\delta_{nm}$. In the case $K < N$ , one refers to these parameters as order parameters and makes the transition from a microscopic to a macroscopic description of the system. However, in our setup such an interpretation is no longer true as we are mainly interested in the case $K>N$. Previous research on initializing the student vectors over a sphere for the RFM with ReLU activations can be found in \cite{Mei2022the,Yehudai2019on,Ghorbani2021linearized} and for a Gaussian (mixture) initialization with quadratic activations in \cite{Ghorbani2019limitation}. \\
The performance of the student with respect to the teacher is measured by the loss function $\epsilon=\frac{1}{2}\left[\zeta-\sigma\right]^2$ known as the mean squared error. As the distribution of the input patterns is accessible, we can take the expectation value of the loss function and define the generalization error $\epsilon_g=\left< \epsilon\left(c_i,\bm{\xi}\right)\right>_{\bm{\xi}}$ depending on the correlation matrices. This makes it possible to analyze the average error of the student evaluated on unseen test data. 
For the error activation function, one finds for the generalization error \cite{saad1995line,goldt2019dynamics}
\begin{align}
\epsilon_{{g,}{\mathrm{erf}}} =\frac{1}{\pi}\Biggl[& \frac{\pi}{6} +\sum_{i,j} {c_i c_j \arcsin\hspace{-2pt}{\left(\frac{Q_{ij}}{2}\right)}} \nonumber \\
 &-2\sum_{i}{ c_i \arcsin\hspace{-2pt}{\left(\frac{R_{i}}{2}\right)}}\Biggr].
 \label{final error reduced erf}
\end{align}
and for a ReLU activation \cite{straat2019line}
\begin{align}
\epsilon_{g, {\mathrm{ReLU}}}  =&\sum_{i,j} \frac{c_ic_j}{2} \biggl[\frac{Q_{ij}}{4} \nonumber \\
&+  \frac{1}{2\pi}\biggl[\sqrt{1-Q_{ij}^2}+Q_{ij}\arcsin\left(Q_{ij}\right)\biggr] \biggr]\nonumber \\
&-\sum_{i} c_i \biggl[\frac{R_{i}}{4} +\frac{1}{2\pi}\biggl[\sqrt{1-R_{i}^2}+R_{i}\arcsin\left( R_{i}\right)\biggr]\biggr]\nonumber \\
&+ \frac{1}{4}.
\label{final error reduced relu}
\end{align}
%with our setup conditions $Q_{ii}=1, T_{nm}=\delta_{nm}$, and 
for the case $M=1$. Here, we have used that $\frac{1}{2}\left<\zeta(\boldsymbol{B},\boldsymbol{\xi})^2\right>_{\mathrm{erf}}=\frac{1}{6}$ and $\frac{1}{2}\left<\zeta(\boldsymbol{B},\boldsymbol{\xi})^2\right>_{\mathrm{ReLU}}=\frac{1}{4}$. \\

Due the special form of the mean squared loss function, the above 
Eqs. (\ref{final error reduced erf}) and (\ref{final error reduced relu}) can be subsumed in the vector-matrix-form 
\begin{align}
\epsilon_g=\frac{1}{2}\left<\zeta(\boldsymbol{B},\boldsymbol{\xi})^2\right>_{\xi}+\frac{1}{2} \bm{c}^T\tilde{\bm{Q}}  \bm{c}- \bm{c}^T  \tilde{\bm{R}}
\label{eps}
\end{align}
with $\tilde{Q}_{ij}=\left<g\left( x_i \right)g\left( x_j \right)\right> $ and $\tilde{R}_{i}=\left<g\left( x_i \right)g\left( y \right)\right> $ describing the correlation between the hidden neurons. We can directly calculate the minimal generalization error by minimizing Eq.~(\ref{eps}) with respect to the $c_i$, and obtain $\epsilon_{g_{\mathrm{min}}}=\frac{\left<\zeta(\boldsymbol{B},\boldsymbol{\xi})^2\right>_{\xi}}{2}-\frac{1}{2}\tilde{\bm{R}}^T \tilde{\bm{Q}}^{-1}\tilde{\bm{R}}.$, where the optimal weights are $\bm{c}_{\mathrm{min}}= \tilde{\bm{Q}}^{-1} \tilde{\bm{R}}$. Therefore, the minimum generalization error depends on the correlations of the hidden neurons and therefore on the choice of the activation function.

\section{Dynamics of the System}
During the learning process, we update the student weights $c_i$ by stochastic gradient descent after each representation of a specific input example according to
\begin{equation}
c_i^{\mu+1}-c_i^{\mu}=-\frac{\eta}{K} \nabla_{c_i} \epsilon \left(c_i^\mu,\xi^\mu\right),
\label{difference eq}
\end{equation}
%In the thermodynamic $N,p,l \rightarrow \infty$ \cite{riegler1995line,goldt2019dynamics,reents+98} and ultra-wide $K \rightarrow \infty$ limit with $\frac{p}{N}=\alpha$, we find a Langevin equation for the student weights
with $\eta$ denoting the learning rate which controls the step size in weight space. The superscript $\mu$ stands for the $\mu-$th input pattern and can be associated with a discreet time measure due to the sequential updating. Commonly, one would rescale the learning rate by $\frac{1}{N}$ in order to study the dynamics of the learning process \cite{Engel2001,riegler1995line,goldt2019dynamics,reents+98}. As in our case the first layer is fixed and just the last $K$ weights are trained, we have rescaled the learning rate by $\frac{1}{K}$ in Eq. (\ref{difference eq}) in order to guarantee small fluctuations for large size limits \cite{Rotskoff2022}. In the ultra-wide limit $N,K,p \rightarrow \infty$ with a finite ratio $\frac{p}{K}=\alpha$, we find a Langevin equation for the student weights
\begin{equation}
\frac{d\bm{c}}{d\alpha}=-\eta\nabla \epsilon_g +  \frac{\eta}{\sqrt{K}} \bm{\gamma}
\label{Langevin}
\end{equation}
where $\bm{\gamma}$ random is a random vector with $\left<\bm{\gamma}\right>=0$, $\left<\gamma_i\left(\alpha\right) \gamma_j\left(\alpha^\prime\right)\right>=\Sigma_{ij} \delta\left(\alpha-\alpha^\prime\right)$ and covariance matrix $\bm{\Sigma}=\left<\left(\nabla \epsilon -\nabla \epsilon_g\right)\left(\nabla \epsilon -\nabla \epsilon_g\right)^T \right>$. \\
In order to derive the Langevin equation given by Eq. (\ref{Langevin}) heuristically, we reconsider the stochastic update rule for a small ratio $\frac{\eta}{K} \rightarrow 0$ and find for the recursion relation
\begin{equation}
\bm{c}^{\mu+l}-\bm{c}^{\mu}\approx-\frac{\eta}{K} l \sum_{j=0}^{l-1} \frac{\nabla \epsilon \left(c^{\mu},\xi^{\mu+j}\right)}{l}.
\end{equation}
Since the input examples are independent identically distributed variables which are presented only once during the whole learning process, we further assume weak correlations in the update steps. Therefore, as the sum on the right hand side is over nearly independent identically distributed variables with a finite variance, we can invoke the Central Limit Theorem \cite{Mandt2017,Latz2021} and approximate the sum by
\begin{equation}
\sum_{j=0}^{l-1} \frac{\nabla \epsilon \left(c^{\mu+j},\xi^{\mu+j}\right)}{l} \approx \nabla \epsilon_g \left(c\right)+\frac{\Delta\bm{\rho}\left(c\right)}{\sqrt{l}},
\label{c difference equation}
\end{equation}
with a random vector $\left<\Delta\bm{\rho}\right>=0$ and corresponding covariance matrix $\textrm{Cov}\hspace{-2pt}\left(\Delta\bm{\rho}\right)=\left<\left(\nabla \epsilon -\nabla \epsilon_g\right)\left(\nabla \epsilon -\nabla \epsilon_g\right)^T \right>$. Due to weak correlations in the update steps, we assume an approximately constant covariance matrix $\Sigma$ with respect to the weights $c_i$. As the covariance matrix is symmetric and positive-semidefinite, we can factorized it $\Sigma=\Omega\Omega^T$ and replace the random vector $\Delta \bm{\rho}$ by $\sqrt{\Delta\alpha}\Delta\bm{\rho} = \Omega \Delta \bm{W}$, where $ \frac{l}{K} = \Delta\alpha$ is a time difference rescaled by the hidden-layer size and $\Delta \bm{W} \in \mathcal{N}\left(0,\bm{1}\right)$ is a Gaussian process. In the ultra-wide limit $K,N,l \rightarrow \infty$, Eq. (\ref{c difference equation}) can be interpreted as a finite-difference equation that approximates the following continuous time stochastic differential equation 
\begin{equation}
d\bm{c}=-\eta\nabla_{c} \epsilon_g d\alpha+\frac{\eta}{\sqrt{K}}\Omega d\bm{W}(\alpha) \ ,
\end{equation}
with $\left<dW_i(\alpha)dW_j(\alpha)\right>=\delta_{ij}d\alpha$ as Brownian motion. The above stochastic differential equation can be recast into a Langevin form by introducing $\left<\bm{\gamma}\right>=0$, $\left<\gamma_i\left(\alpha\right) \gamma_j\left(\alpha^\prime\right)\right>=\Sigma_{ij} \delta\left(\alpha-\alpha^\prime\right)$ and $\Omega dW(\alpha) = \int_{0}^{d\alpha}d\tau \bm{\gamma}\left(\tau\right)$. More details for the Langevin equivalence of the sotchastic gradient descent can be found in \cite{Hu2019,Mandt2017,Latz2021}.\\
\begingroup
\centering
\begin{figure}[t]
\centering
\subfigure{\includegraphics[width=0.75\linewidth]{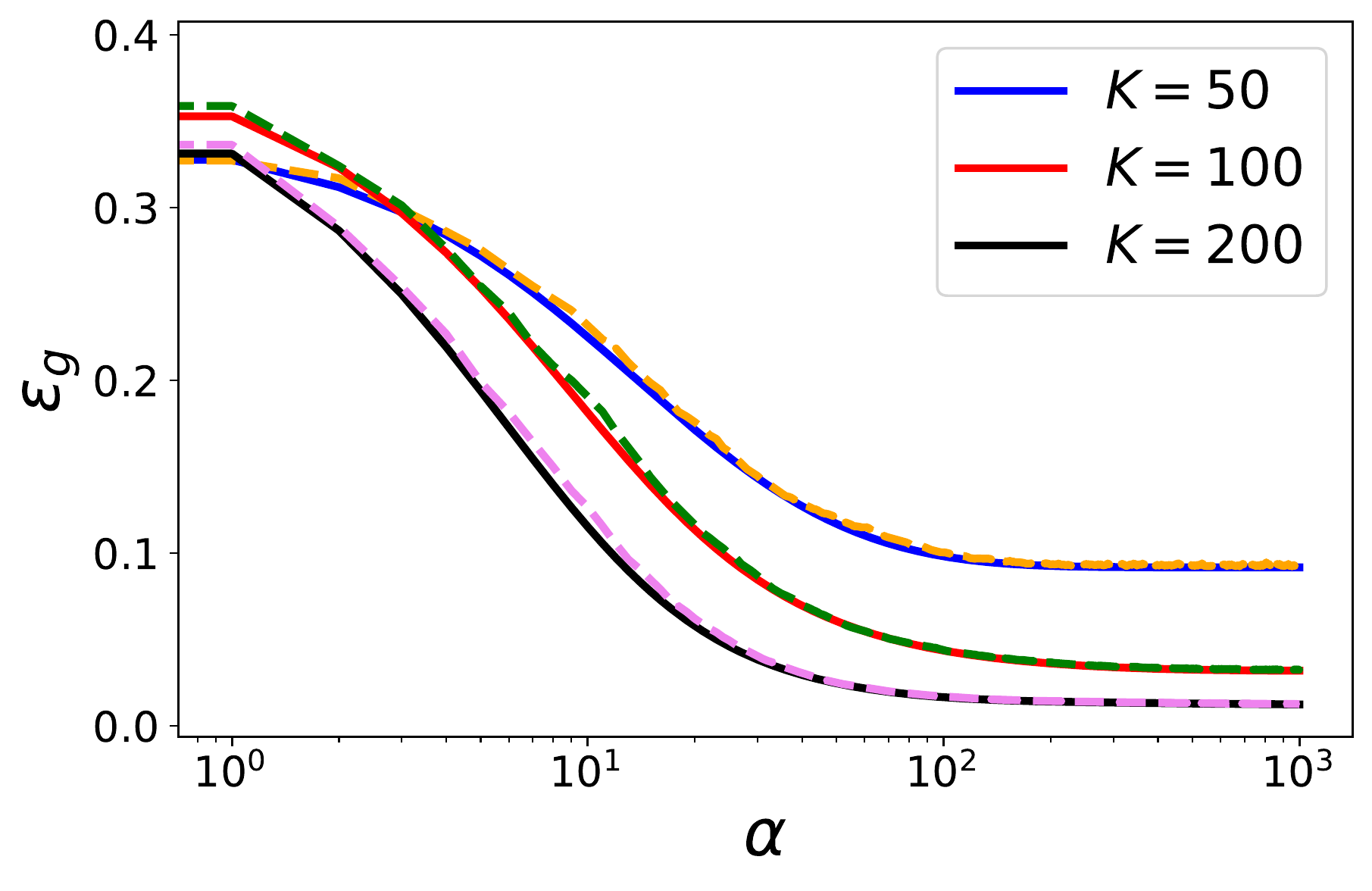}}\label{fig:1a}\\
\subfigure {\includegraphics[width=0.75\linewidth]{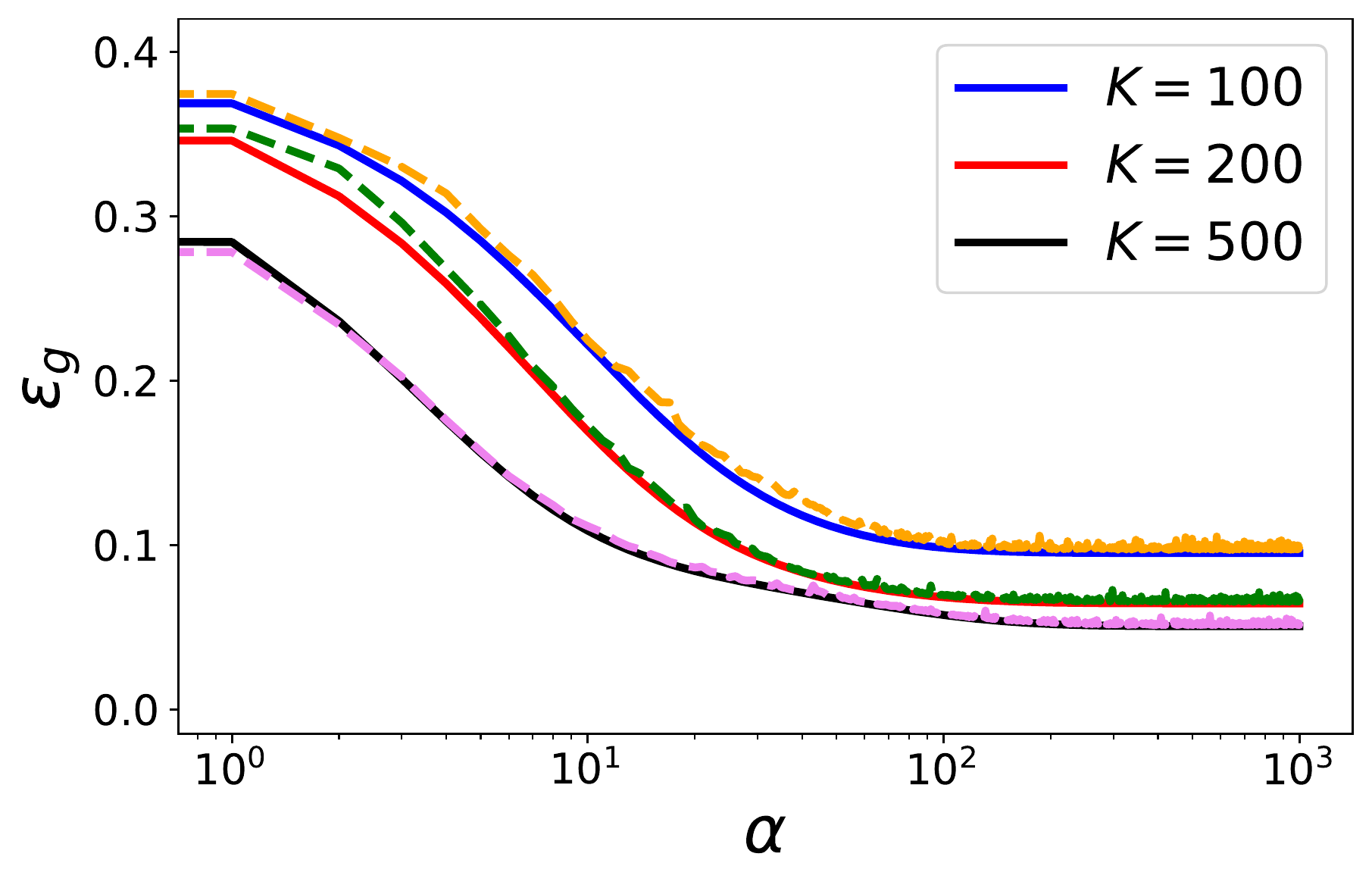}}\label{fig:1b}
\caption{Generalization error as a function of $\alpha$ for an error function activation $g\hspace{-2pt}\left(x\right)=\mathrm{erf}\hspace{-2pt}\left(\frac{x}{\sqrt{2}}\right)$ (top) and ReLU activation function (bottom) for $N=100$. The numerical solutions of the differential equations (solid lines) fits well to the simulations. We use for the simulations $\eta_{\mathrm{simulation}}=\frac{0.1}{K}$ (dashed lines) corresponding to the rescaling of the learning rate in Eq. (\ref{difference eq}), and for the solutions of the differential equations given by Eq. (\ref{mean path}) $\eta=0.1$. For both methods, we set the same values for $\boldsymbol{R},\boldsymbol{Q}$ and use the same initial values of $\boldsymbol{c}_0\in\mathcal{N}\left(0,\frac{1}{K}\right)$. } \label{fig: simulation vs odes}
\end{figure}
\endgroup
In order to make further assertions about the large-$K$ limit on the right-hand side of Eq.~(\ref{Langevin}), we need to take a closer look at the variance of the trajectory. As the system size increases, one can replace the above stochastic Langevin equation by its mean trajectory leading to a deterministic differential equation if the fluctuations become negligible \cite{VANKAMPEN2007}. As a measure for the fluctuations, we consider the relative variance of the stochastic trajectory and find 
\begin{align}
\frac{\left<\left(\frac{d\bm{c}}{d\alpha}\right)^2\right>-\left<\frac{d\bm{c}}{d\alpha}\right>^2}{\left<\frac{d\bm{c}}{d\alpha}\right>^2} &= \frac{\eta^2}{K}\frac{\left<(\nabla\epsilon)^2\right>-(\nabla\epsilon_g)^2}{(\nabla\epsilon_g)^2}  \nonumber \\
& \propto \frac{1}{K}.
\label{fluctuations}
\end{align}
directly related to the fluctuations of the loss function, which are of order unity. 
%In the Supplemental Material, we derive the scaling of the relative variance given in Equation (\ref{fluctuations}). 
Thus, as the system size increases, the relative variance of the loss function converges to zero and we can replace the stochastic Langevin equation by its mean in the ultra-wide limit. In statistical mechanics, such a scaling relation for the variance of a system's property is known as a self-averaging character. A rigorous treatment of the fluctuations of the stochastic gradient descent can be found in \cite{Rotskoff2022}. \\
We find for the mean path
\begin{align}
\left<\frac{d\bm{c}}{d\alpha}\right>=-\eta \left[\tilde{\bm{Q}}\bm{c}-\tilde{\bm{R}}\right],
\label{mean path}
\end{align}
depending on the choice of the activation function. Therefore, we obtain the set of deterministic differential equations  Eq.~(\ref{mean path}) characterizing the dynamical behavior of the learning process that is valid for arbitrary ratios $\frac{K}{N}$. The fixed point of Eq.~(\ref{mean path}) reveals the asymptotic solutions for the weights
\begin{equation}
\bm{c}^*= \tilde{\bm{Q}}^{-1} \tilde{\bm{R}} \ ,
\label{asymp weights}
\end{equation}
and for the generalization error
\begin{equation}
\epsilon_g^*=\frac{\left<\zeta(\boldsymbol{B},\boldsymbol{\xi})^2\right>_{\xi}}{2}-\frac{1}{2}\tilde{\bm{R}}^T \tilde{\bm{Q}}^{-1}\tilde{\bm{R}} \ .
\label{asym eg}
\end{equation}
As one sees immediately, the asymptotic solution corresponds to the minimal generalization error meaning that the error will indeed be minimized after a sufficiently long training period. Figure (\ref{fig: simulation vs odes}) compares the generalization error of the random feature model for simulations according to the update rule given by Eq. (\ref{difference eq}) with the solutions of Eq. (\ref{mean path}). 

\section{Asymptotic Generalization Error}
In order to analyze how the asymptotic generalization error depends on the student size $K$, we first consider a finite hidden-to-input layer size ratio $K=\beta N$ with $\beta \in (1,\infty)$. In the ultra-wide limit, we can linearize $\tilde{\bm{Q}}$ and $\tilde{\bm{R}}$ to first order in $R_i$ and $Q_{ij}$ since large overlaps are unlikely to occur for a finite $\beta$. This assumption is based on the curse of dimensionality where we consider $K$ random $N$-dimensional student vectors leading to small overlaps of $O(1/\sqrt{N})$. Furthermore, the generalization error depends on the distribution of the student and teacher vectors and we therefore analyze the dynamics of the typical asymptotic generalization by taking the expectation value of Eq. (\ref{asym eg}) in this linearized regime.\\ 
First, we present the calculations for the error function activation. After evaluating Eq.~(\ref{final error reduced erf}) for the fixed point solution in the linearized regime, we rewrite it in a compact form 
\begin{equation}
\epsilon_g^*=\frac{1}{2\pi}\left(\frac{\pi}{3}-\bm{R}^T \bm{S}^{-1}\bm{R}\right)
\label{eps linear erf}
\end{equation}
with $S_{ii}=\frac{\pi}{3}$ and $S_{ij}=Q_{ij}$. We use the definition of the correlation matrices to evaluate the expectation value of the right hand side of Eq.~(\ref{eps linear erf}) as
\begin{align}
&\left<\bm{R}^T \bm{S}^{-1}\bm{R}\right>_{Q_{ij},R_i} \nonumber \\
&= \frac{1}{N^2}\sum_{ij}^K \sum_{ab}^N\left< J_{ia}B_a\left(S^{-1}\right)_{ij}J_{jb}B_b\right>_{\bm{J}_{i},\bm{B}} \nonumber \\
&=\frac{1}{N}\sum_{ij}^K \sum_{ab}^N\left< J_{ia}J_{ja}\left(S^{-1}\right)_{ij}\right>_{\bm{J}_{i}},
\end{align}
where we have exploited $\left<B_aB_b\right>=\delta_{ab}$ to obtain the last equation. Next, we use the definition of the trace of a product of matrices in order to get
\begin{align}
\left<\bm{R}^T \bm{S}^{-1}\bm{R}\right>&=\frac{1}{N} \left< \text{tr}\left(\bm{Q} \bm{S}^{-1}\right)\right> \nonumber \\
&=\frac{1}{N} \sum_{i}^K \left< \frac{\lambda_i}{\frac{\pi}{3}-1+\lambda_i}\right>,
\label{expect}
\end{align}
with $\lambda_i$ as the eigenvalues of the random matrix $\bm{Q}$. Since $\bm{J}$ is a $K \times N$ random matrix whose entries have zero mean and bounded variance, the eigenvalues of the correlation matrix for $N \rightarrow \infty $ are distributed according to the Mar{\v{c}}enko-Pastur distribution \cite{Marenko1967}. For ratios $\frac{K}{N}>1$ the Mar{\v{c}}enko-Pastur distribution consists of two parts: the first $K-N$ eigenvalues are zero and just $N$ eigenvalues contribute to the sum in Eq. (\ref{expect}). For the remaining expectation value, we insert Mar{\v{c}}enko-Pastur law for the eigenvalue distribution
\begin{align}
&\frac{1}{N}\sum_{l}^N\left< \frac{\lambda_l}{\frac{\pi}{3}-1+\lambda_l}\right> \nonumber \\
& =\frac{1}{2\pi N} \sum_{l}^N \int_{\lambda_-}^{\lambda_+} \frac{\sqrt{\left(\lambda_l-\lambda_-\right)\left(\lambda_+-\lambda_l\right)}}{\frac{\pi}{3}-1+\lambda_l} d\lambda_l,\nonumber \\
&=\frac{1}{4}\left[\lambda_+ + \lambda_-+2c+2 \sqrt{\lambda_+\lambda_-+\left(\lambda_+ + \lambda_-\right)c+c^2}\right],
\end{align}
with $\lambda_{\pm}=\left(1\pm\sqrt{\frac{K}{N}}\right)^2$ and $c=\frac{\pi}{3}-1$. After inserting the definition of $\lambda_{\pm}$, we finally obtain
\begin{align}
\left<\bm{R}^T \bm{S}^{-1}\bm{R}\right> & = \frac{1}{2} \left[  \frac{K}{N}+\frac{\pi}{3} \vphantom{\sqrt{\left(c+\left(\frac{K}{N}+1\right)\right)^2-\frac{4K}{N}} } \right. \nonumber \\
  &\left. \quad - \sqrt{\left(c+\left(\frac{K}{N}+1\right)\right)^2-\frac{4K}{N}} \right] \ .
  \label{exact sol}
\end{align}
By a first-order Taylor expansion in $\frac{N}{K}$, we find $\lim_{\frac{N}{K} \rightarrow 0 }\lim_{K,N \rightarrow \infty}\left<\bm{R}^T \bm{S}^{-1}\bm{R}\right>=1$. This result leads to 
\begin{equation}
\lim_{\frac{N}{K} \rightarrow 0 }\lim_{K,N \rightarrow \infty}\left<\epsilon_{g_{\mathrm{erf}}}^*\right>=\frac{1}{2\pi}\left(\frac{\pi}{3}-1\right) \approx 0.0075
\label{plateau}
\end{equation}
as a limiting value of the asymptotic generalization error in the linearized regime.\\
\begingroup
\centering
\begin{figure}[t]
\centering
\subfigure{\includegraphics[width=0.75\linewidth]{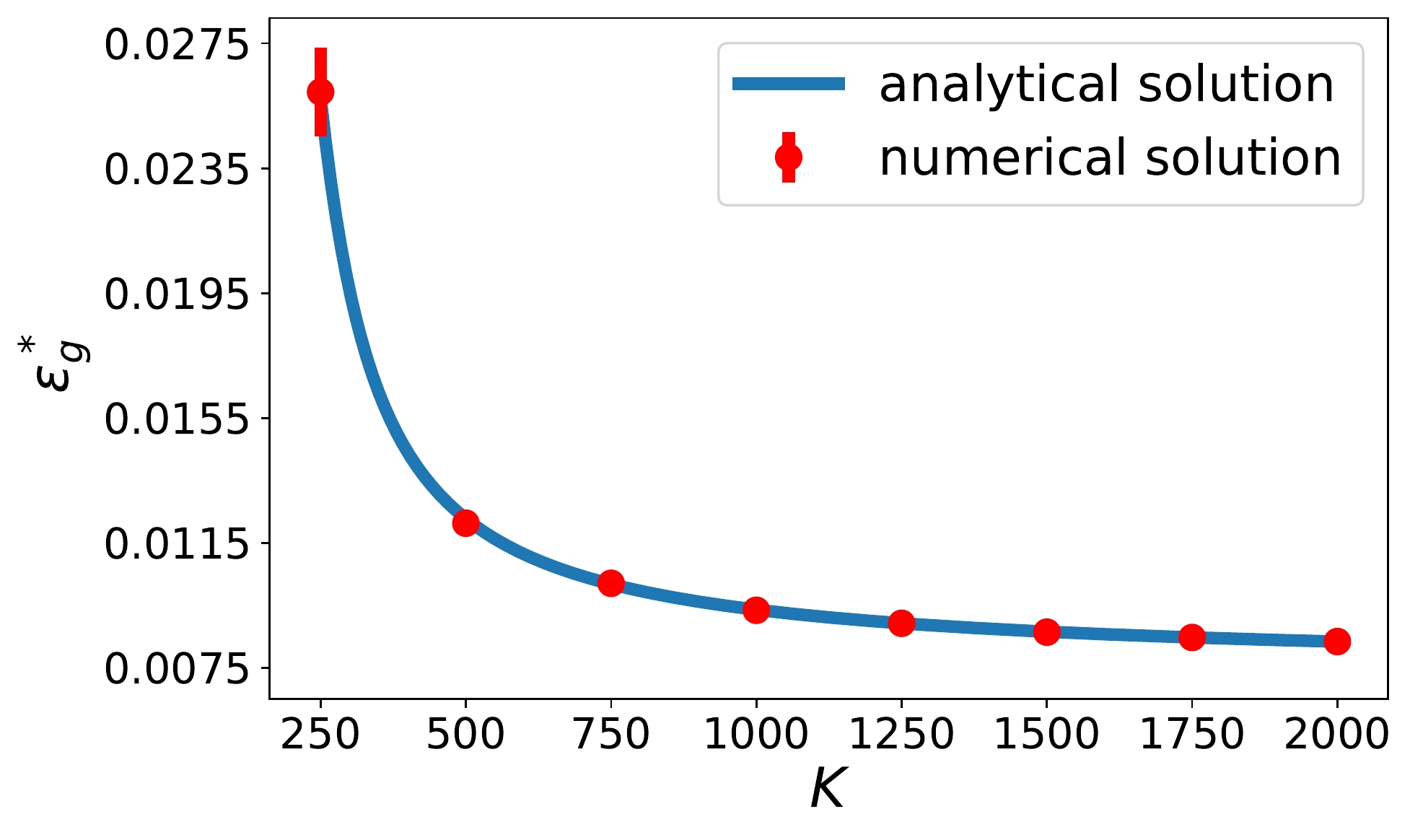}}\label{fig:2a}\\
\subfigure {\includegraphics[width=0.75\linewidth]{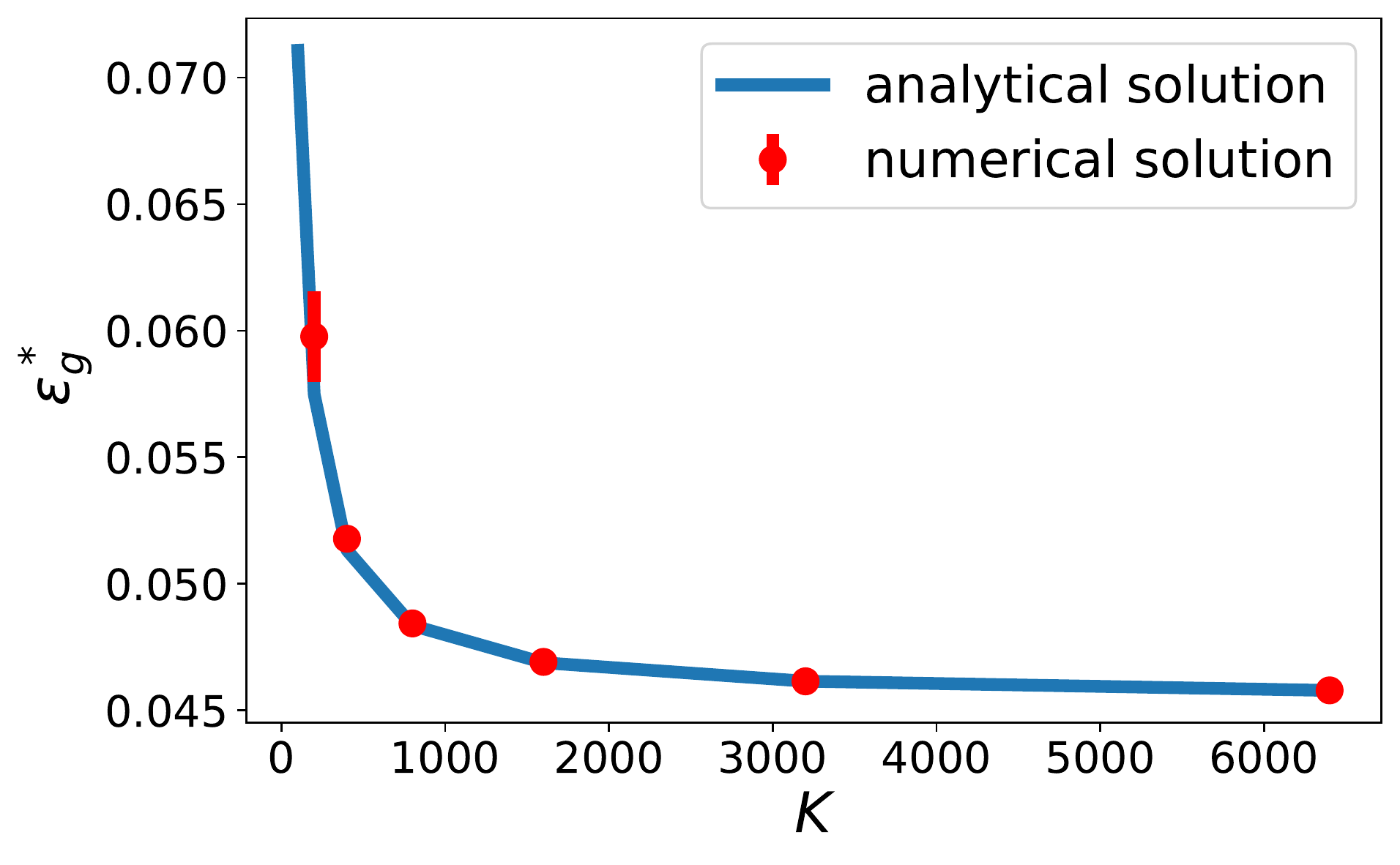}}\label{fig:2b}
\caption{ Asymptotic generalization error $\epsilon_g^*$ as a function of $K$ for $ M = 1$ for a error function as the activation with $N = 200$ (top) and ReLU activation with $N=50$ (bottom). The numerical solution of Eq. (\ref{asym eg}) is obtained from ten initializations of random matrices $\bm{R}$ and $\bm{Q}$ under the linearized setup. The errorbars show the standard deviation of these averages. The analytical solution is based on Eq. (\ref{exact sol}) for the error function and on Eq. (\ref{exact sol}) for the ReLU activation.}
\label{fig: eg von K}
\end{figure}
\endgroup
Next, we evaluate the asymptotic generalization error for the ReLU activation function in the linearized regime. For this we linearize Eq. (\ref{final error reduced relu}) and rewrite this again in a compact form
\begin{equation}
\epsilon_g^*=\frac{1}{4}-\frac{1}{2}\hat{\bm{R}}^T \hat{\bm{Q}}^{-1}\hat{\bm{R}}
\label{eps linear relu}
\end{equation}
with $\hat{Q}_{ii}=\frac{1}{2}$, $\hat{Q}_{ij}=Q_{ij}+\frac{1}{2\pi}$ and $\hat{R}_i=\frac{1}{4}R_i+\frac{1}{2\pi}$. As for the error function case, we take the expectation value of the r.h.s of Eq. (\ref{eps linear relu}) and use the definition of $\hat{\bm{R}}$
\begin{align}
\left<\hat{\bm{R}}^T \hat{\bm{Q}}^{-1}\hat{\bm{R}}\right>&= \frac{1}{16} \left<\bm{R}^T \hat{\bm{Q}}^{-1}\bm{R}\right> \nonumber \\
&+\frac{1}{8\pi}\left[\left<\bm{R}^T \bm{T}\right>+\left< \bm{T}^T\bm{R}\right>\right] \nonumber \\
&+\frac{1}{4\pi^2}\sum_{ij}\left(\hat{\bm{Q}}^{-1}\right)_{ij},
\label{expec relu terms}
\end{align}
where $\bm{T}^T$ is a vector containing the sum of the columns of $\hat{\bm{Q}}^{-1}$, i.e. $T_i=\sum_j \left(\hat{\bm{Q}}^{-1}\right)_{ij}$. The second term of Eq.~(\ref{expec relu terms}) vanishes as $\bm{R}$ occurs only linearly and $\left<\bm{R}\right>=0$. After exploiting the definition of $\bm{R}$ and that the entries of the teacher vector are i.i.d, we obtain for the first term of Eq.~(\ref{expec relu terms})
\begin{align}
\left<\bm{R}^T \hat{\bm{Q}}^{-1}\bm{R}\right>=\frac{1}{N} \left< \text{tr}\left(\bm{Q} \hat{\bm{Q}}^{-1}\right)\right> .
\label{expect_relu}
\end{align}
With the help of the Sherman-Morrison formula, we can calculate $\hat{\bm{Q}}^{-1}$ and find
\begin{equation}
\hat{\bm{Q}}^{-1}=\bm{A}^{-1} - \frac{\bm{A}^{-1} v v^T \bm{A}^{-1}}{2 \pi \left(1+\frac{1}{2\pi} v^T \bm{A}^{-1} v\right)}
\label{Sherman Morrison formula}
\end{equation}
with $\bm{A}=\frac{1}{4}\left(\bm{Q}+\mathbb{I}-\frac{2}{\pi}\mathbb{I}\right), v=\left(1,1,...,1\right)^T$ and $\mathbb{I}$ as the identity matrix. For the first part of the trace, we obtain with the same arguments as for the error function activation $\frac{1}{N} \left< \text{tr}\left(\bm{Q}\bm{A}^{-1}\right)\right> =\frac{1}{N} \sum_{i}^K \left< \frac{4\lambda_i}{\lambda_i+1-\frac{2}{\pi}}\right> $ and in the ultra-wide limit $ \lim_{\frac{N}{K} \rightarrow 0} \lim_{K,N \rightarrow \infty}  \frac{1}{N} \sum_{i}^K \left< \frac{4\lambda_i}{\lambda_i+1-\frac{2}{\pi}}\right>=4 \hspace{2pt}$. As shown in the Appendix, we find for the second part $\lim_{K,N \rightarrow \infty} \left< \frac{1}{N}\text{tr}\left(\frac{\bm{Q} \bm{A}^{-1} v v^T \bm{A}^{-1}}{2 \pi \left(1+\frac{1}{2\pi} v^T \bm{A}^{-1} v\right)}\right)\right> \propto \frac{1}{N} \hspace{2pt}$. Thus, we conclude
\begin{equation}
\lim_{\frac{N}{K} \rightarrow 0} \lim_{K,N \rightarrow \infty}  \left<\bm{R}^T \hat{\bm{Q}}^{-1}\bm{R}\right>=4.
\label{part1}
\end{equation}
Now, we evaluate the third term given in Eq.~(\ref{expec relu terms}). For this, we sum over the entries of $\hat{\bm{Q}}^{-1}$ with the help of Eq.~(\ref{Sherman Morrison formula}). In the thermodynamic limit $N\rightarrow \infty$, we can assume $Q_{ij}+\frac{1}{2\pi}\approx\frac{1}{2\pi}$ since $Q_{ij} \propto \frac{1}{\sqrt{N}}$. Therefore, we obtain in this limit $\bm{A}\approx\left(\frac{1}{2}-\frac{1}{2\pi}\right)\mathbb{I} $ and for the sum
\begin{align}
 \sum_{ij}^K\left(\hat{\bm{Q}}^{-1}\right)_{ij}&= Ka \left(1- \frac{Ka}{2\pi}\frac{1}{1+\frac{Ka}{2\pi}} \right) -\mathcal{O}\hspace{-2pt}\left(\frac{N}{K}\right)  \nonumber \\
&\overset{\frac{N}{K} \rightarrow 0}{ \underset{K,N \rightarrow \infty}{=}} 2\pi.
\label{part2}
\end{align}
with $a=\frac{1}{\frac{1}{2}-\frac{1}{2\pi}}$. After inserting Eqs.~(\ref{part1}) and (\ref{part2}) into Eq.~(\ref{eps linear relu}), we obtain for the asymptotic generalization error in the linearized regime
\begin{equation}
\lim_{\frac{N}{K} \rightarrow 0 }\lim_{K,N \rightarrow \infty} \left<\epsilon_{g_{\mathrm{ReLU}}}^*\right>=\frac{1}{8}-\frac{1}{4\pi}\approx 0.045 
\label{plateau relu}
\end{equation}
\begingroup
\centering
\begin{figure}[t]
\centering
\subfigure{\includegraphics[width=0.75\linewidth]{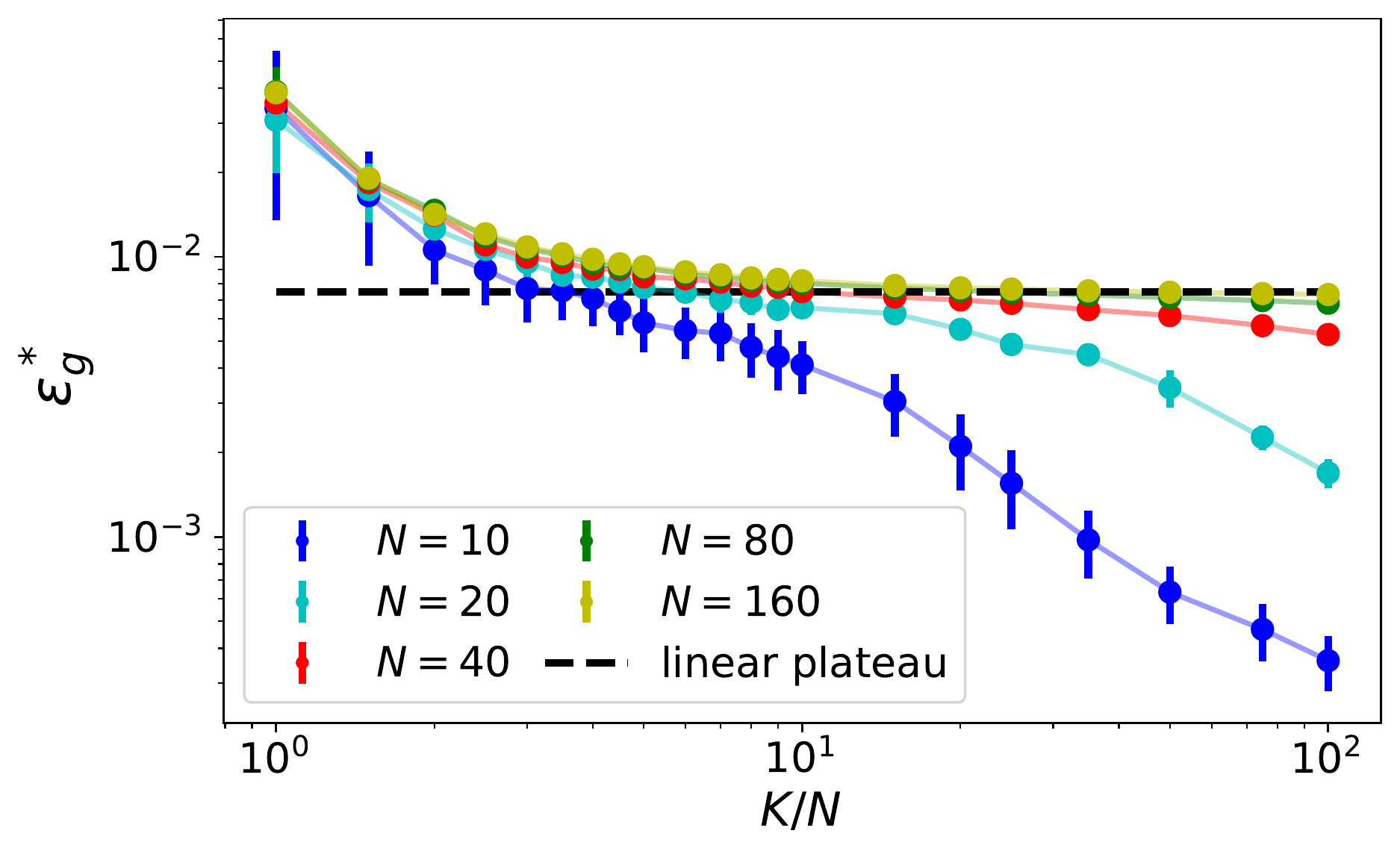}}\label{fig:3a}\\
\subfigure {\includegraphics[width=0.75\linewidth]{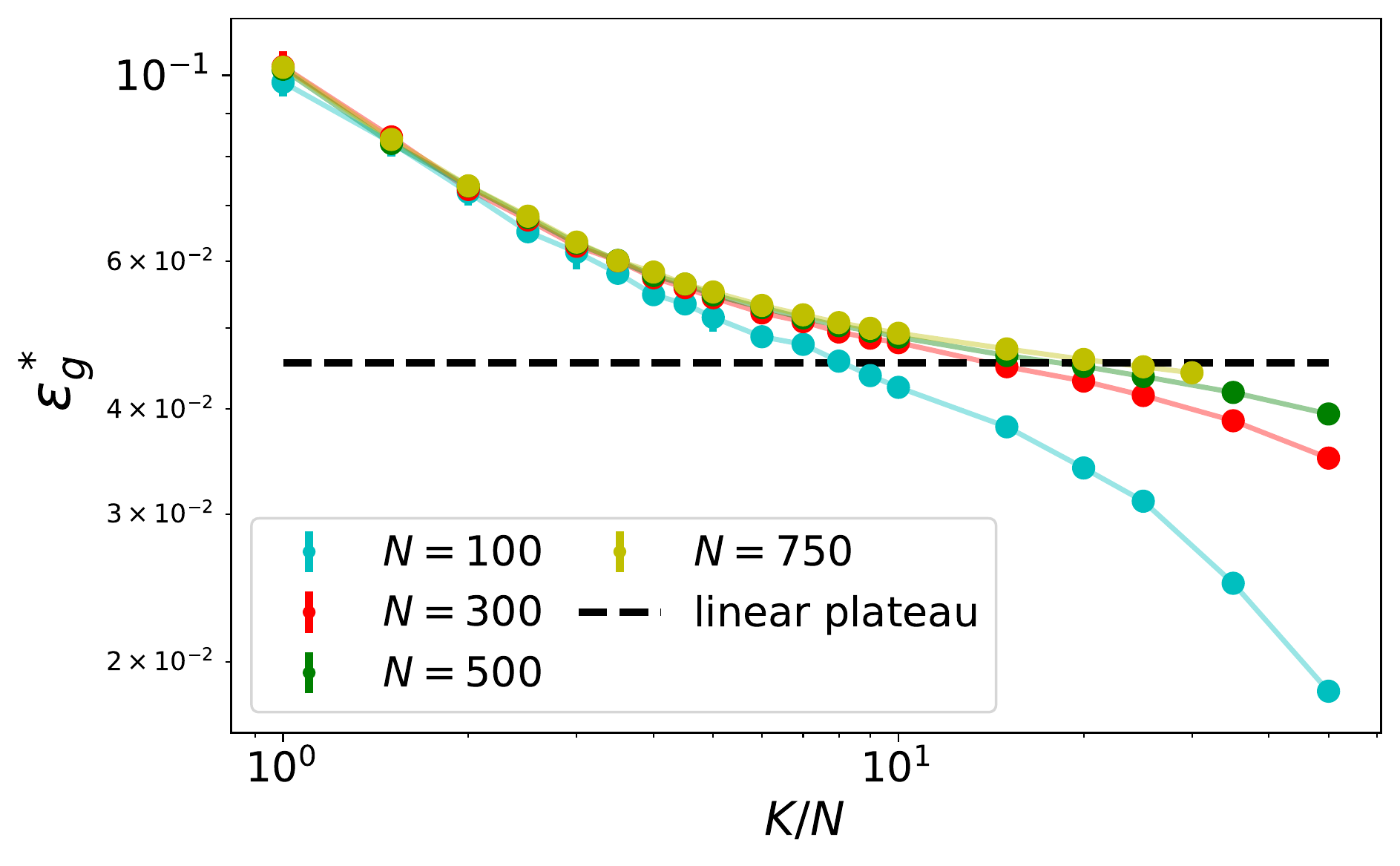}}\label{fig:3b}
\caption{Asymptotic generalization error as a function of $\frac{K}{N}$ for an error activation $g\hspace{-2pt}\left(x\right)=\mathrm{erf}\hspace{-2pt}\left(\frac{x}{\sqrt{2}}\right)$ (top) and ReLU (bottom). Here, we evaluate Eq.~(\ref{asym eg}) numericall and compute the expectation value of the generalization error for a given $\frac{K}{N}$. The errorbars show the standard deviation of the average over $100$ initializations for $N=10$ and $10$ initializations for $N>10$. The asymptotic generalization error increases with $N$, even though the hidden layer size increases proportional to that of the input layer. For large enough $N$ and finite $\frac{K}{N}$, the generalization error obtains the limit value for the linearized scenario given in Eq. (\ref{plateau}) for the error function and Eq. (\ref{plateau relu}) for the ReLU activation function.} \label{fig: eg von K/N}
\end{figure}
\endgroup
Figure (\ref{fig: eg von K}) shows the asymptotic generalization error given by Eq.~(\ref{exact sol}) together the corresponding numerical solution of the expectation value averaged over different initializations of $\bm{R}$ and $\bm{Q}$ in the linearized setting.
Figure (\ref{fig: eg von K/N}) displays the numerical solution of Eq.~(\ref{asym eg}) for the error function averaged over several matrix initializations not restricted to the linearized setting and shows how the linearized regime is reached in regard of the ratio $\frac{K}{N}$. For small ratios, we are in the linearized regime for which we have small correlations between the student and teacher vectors whereas for large enough ratios, one obtains a transition where linearization is no longer valid. However, the point of transition is different for different ratios depending on the input dimension $N$ and is shifted to the right as $N$ increases. Therefore, we conclude that the linearized regime is indeed valid for $N,K \rightarrow \infty$ under a fixed and finite ratio.
\begin{figure}[t]
\centering
\includegraphics[width=1\linewidth]{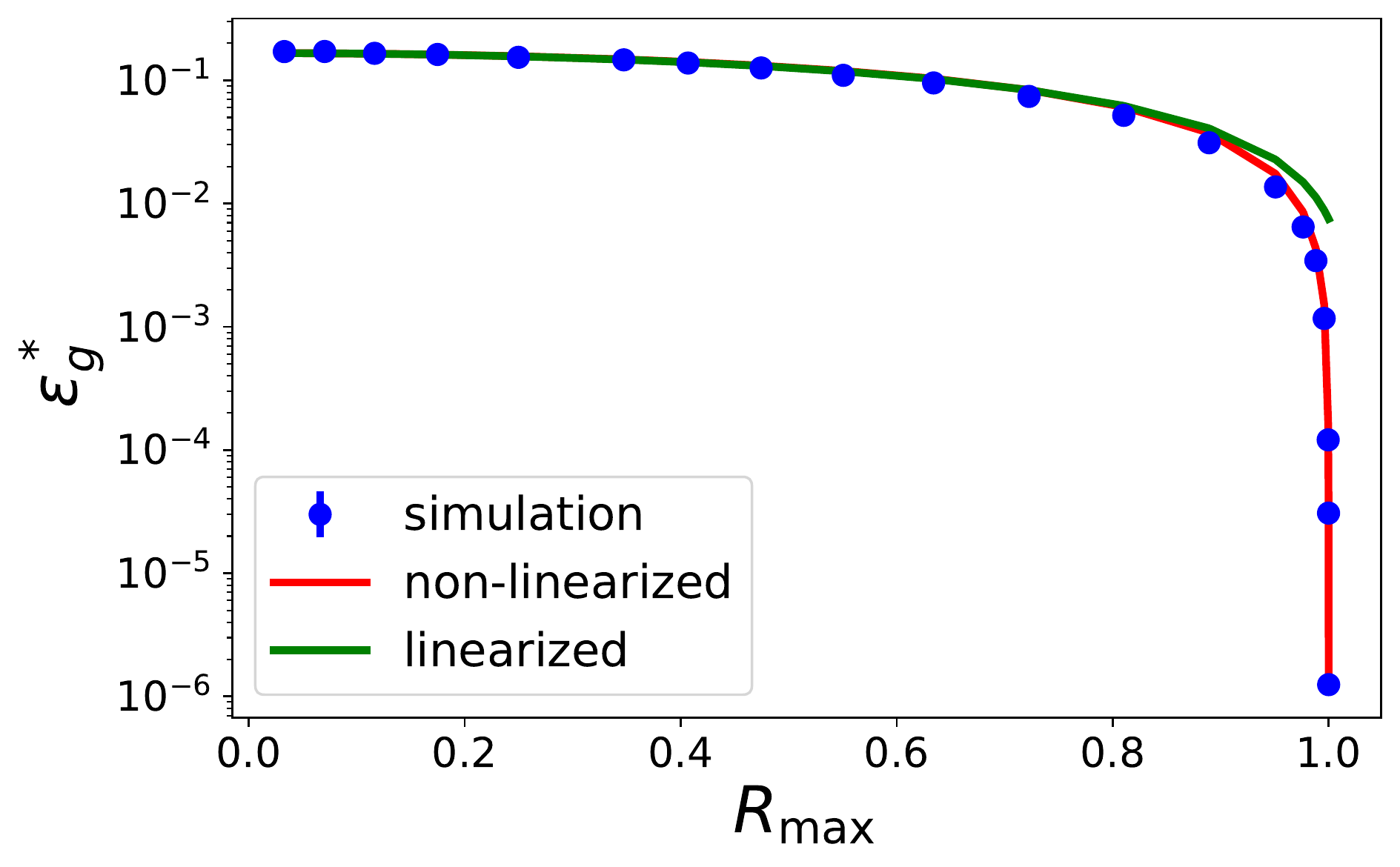}
\caption{ Asymptotic generalization error as a function of $R_{\mathrm{max}}=\max_{1 \leq i \leq K}\hspace{2pt} R \in \{R_i\}$ for $K=7$, $M=1$, $N=5$ and $\eta=\frac{0.1}{K}$ for a error function as the activation $g\hspace{-2pt}\left(x\right)=\mathrm{erf}\hspace{-2pt}\left(\frac{x}{\sqrt{2}}\right)$. We initialize the student and set the teacher in such a way that the first component of $\bm{R}$ is the largest one and the other are small and of similar size. The blue curve shows the corresponding simulation. The orange and green curves show the solution of Eq. (\ref{asym eg}) for a perceptron and a linearized perceptron, respectively. For the simulation, we averaged the generalization error over a predefined interval to get its asymptotic value. The errorbars (standard deviation of this average) are smaller than the symbol size. }
%However, the small $R_{i>1}$ have a significant contribution for the generalization and therefore the blue curve lies lower than the orange one. 
\label{fig: at least on R}
\end{figure}

\section{Perfect Learning Scenario}
Numerically, it turns out that for a constant $N$ and increasing $K$ the asymptotic generalization error decreases towards zero (cf. Figure (\ref{fig: eg von K/N})). However, for a constant ratio $\frac{K}{N}$ the generalization error increases as the input size $N$ becomes larger meaning that the student size $K$ is not big enough to maintain a small generalization error. The reason for that lies in the probability distribution of the overlaps $R_i$. The larger the student size $K$ becomes, the higher is the probability to pick a student vector that has a high overlap with the teacher vector under a constant input size $N$. The remaining problem is to figure out how fast the student size $K$ must grow with $N$ to obtain a small asymptotic generalization error. Accompanying this, the more student vectors are approximately in the direction of the teaching vector, the smaller the generalization error is in general, making a linearization of $\hat{\bm{R}}$ and $ \hat{\bm{Q}}$ infeasible. Moreover, if only one student vector shows a high overlap, then the generalization error already decreases towards zero as $R_i \rightarrow 1$ (cf. Figure (\ref{fig: at least on R})). \\
Therefore, we ask for the probability to find at least one large overlap $\mathrm{max} \{R_i\} > R^*$ after the initialization of $K$ student vectors for a given $N$ and threshold $R^*$. We use the relation
\begin{equation}
P\left(\mathrm{max} \{R_i\} > R^*;N,K\right)=1-F\left(R^*;N\right)^K,
\label{simple ansatz}
\end{equation}
where $F\left(R^*;N\right)= \mathrm{Pr}\left(R \leq R^* ;N \right)=\int_{0}^{R^*} \hspace{-2pt}dR_i \hspace{2pt}\rho\hspace{-2pt}\left(R_i\right)$ is the cumulative probability to find $R \leq R^*$ after randomly drawing a student vector, obtained by integrating over a density function $\rho\hspace{-2pt}\left(R\right)$. Since the student and teacher vectors are drawn from a uniform distribution over the $N$-sphere, the shifted overlaps $\frac{R_i+1}{2}$ are generated by the beta distribution. For the error function activation, we obtain for the density $\rho\left(y\right)=\lvert\cos\left(y\right)\rvert \beta\left(\frac{2\sin\left(y\right)+1}{2};a,b\right)$, where $\beta\left(t;a,b\right)=\frac{1}{B(a,b)} t^{a-1}(1-t)^{b-1}$ is the beta density function with normalization constant $B(a,b)=\int_0^{1} t^{a-1}(1-t)^{b-1}\hspace{3pt}dt$ and $a=b=\frac{N-1}{2}$. We can now perform the integration for $R^*>R$, thus obtaining the probability for large overlaps between student and teacher vectors
\begin{align}
F\left(R^*;N\right)& =1-\int_{\arcsin\left(\frac{R^*}{2}\right)}^{\frac{\pi}{6}} \rho\left(y\right) \hspace{3pt}dy \nonumber \\
&=\frac{B\left(\frac{R^*+1}{2},a,a\right)}{B\left(a,a\right)}\nonumber \\
&= I_z\left(a,a\right) \ . 
\label{ansatz}
\end{align}
%Thus, we find $F\left(R^*;N\right)\geq 1-\exp\left(-\frac{R^{*^2} N}{2}\right)$.
Here, $B(z,\alpha,\beta)=\int_0^{z} t^{\alpha-1}(1-t)^{\beta-1}\hspace{3pt}dt$ is the incomplete beta function with $z=\frac{R^*+1}{2}$ for the case above and $I_z\left(a,b\right)$ is the regularized incomplete beta function. Therefore, our cumulative distribution function is related to the Binomial cumulative distribution function via the regularized incomplete beta function $1-F\left(R^*;N\right)=F_{\mathrm{Binomial}}\left(\frac{N-1}{2};N-2,\frac{R^*+1}{2}\right)$. For the case ${R^*} \lesssim 1$, one can estimate the tail bounds of the Binomial distribution function by the Chernoff bound $F_{\mathrm{Binomial}}\left(\frac{N-1}{2};N-2,\frac{R^*+1}{2}\right) \geq \frac{1}{\sqrt{2N-4}}\exp\left[-\left(N-2\right)D\left(\frac{N-3}{2N-4} \bigg \Vert \frac{R*+1}{2}\right)\right]$ with the Kullback-Leibler divergence $D\left( P \Vert Q\right)= P \ln\left(\frac{P}{Q}\right) +(1-P)\log\left(\frac{1-P}{1-Q}\right)$ \cite{robert1990}. \\
Furthermore, we demand $P\left(\mathrm{max} \{R_i\} > R^*;N,K\right) >P^*$, where $P^*$ is a probability threshold or confidence and insert this condition in Eq. (\ref{simple ansatz}) which yields $K=\frac{\ln\left(1-P^*\right)}{\ln\left(F\left(R^*;N\right)\right)}$. Finally, we obtain
\begin{equation}
K > \sqrt{2N-4} \hspace{5pt} e^{\frac{N}{2} \ln\left(\frac{1}{1-{R^*}^2}\right)  } \hspace{3pt} \lvert  \ln\left(1-P^*\right)\rvert \hspace{2pt}.
\label{K_depen}
\end{equation}
Therefore, the student size $K$ has to increase exponentially fast with the input-layer size $N$ for a fixed $R^*$ and $P^*$ leading to exponetial long training times if one wants to keep a small generalizaton error below the threshold of the linearized regime. This result is consistent with the conclusions in \cite{Yehudai2019on}.

\section{Conclusion}
In conclusion, we have studied the learning dynamics of the random feature model trained by the stochastic gradient descent embedded in the student-teacher framework. We obtained asymptotic solutions of the generalization error out of a set of coupled differential equations describing the weight dynamics. For a regime with a finite ratio of hidden layer width and input dimension, we computed the asymptotic generalization error and found that it stays finite for two choices of activation functions. In the second part of this work, we found by a simple ansatz that the generalization error can become arbitrarily small under an exponential increase of the student size in relation to the input dimension.

\bibliography{Random_Feature_Model.bib}
\bibliographystyle{icml2023}

\section{Appendix}
\label{Appendix}
The code for this work can be find \href{https://github.com/ICML23AnonymousRFM/Random-Feature-Model}{ here}.

\subsection{Exact expressions for the system}
In order to derive the exact expression for the generalization error $\epsilon_g=\left< \epsilon\left(c_i,\bm{\xi}\right)\right>_{\bm{\xi}}$ and the differential equations, we introduce local fields $x_i^\mu =\frac{\bm{J}_i^\mu \bm{\xi^\mu }}{\sqrt{N}}, y_n^\mu =\frac{\bm{B}_n\bm{\xi}^\mu }{\sqrt{N}}$. These local fields describe the overlaps between the student and teacher vectors with the input vector. Due to the special form of the mean squared loss function, we obtain the following structure for the generalization error
\begin{align}
\epsilon_g =& \frac{1}{2} \sum_{i,j}^K c_i c_j I_2(i,j) + \frac{1}{2M} \sum_{n,m}^M I_2(n,m) \nonumber \\
& -\frac{1}{\sqrt{M}} \sum_{i}^K\sum_{n}^M c_i I_2(i,n)
\end{align}
with $I_2(i,j)=\tilde{\bm{Q}} =\left<g\left( x_i \right)g\left( x_j \right)\right>_{\xi}, I_2(i,n)= \tilde{\bm{R}}=\left<g\left( x_i \right)g\left( y_n \right)\right>_{\xi}$. Therefore, the exact generalization error depends on the choice of the activation function. \\
The expressions $I_2$ are integrals over the local fields which are solved in closed form for error activation functions by \cite{saad1995line,goldt2019dynamics} and for a ReLU function in \cite{straat2019line} for the soft committee machine. For example, one finds for the error activation function a $2\times 2-$ sub-covariance matrix, which contains the relevant terms
\begin{equation}\label{I2}
    I_2=\frac{2}{\pi}\;\arcsin\left( \frac{C_{12}}{\sqrt{1+C_{11}}\sqrt{1+C_{22}}}\right)
\end{equation}
and for the ReLU function
\begin{align}
    I_2=&\frac{C_{12}}{4}+ \nonumber \\ 
& \frac{1}{2\pi}\biggl[\sqrt{C_{11}C_{12}-C_{12}^2}+C_{12}\arcsin\left( \frac{C_{12}}{\sqrt{C_{11}C_{22}}}\right)\biggr].
\end{align}
For example, one obtains the following sub-matrix for the student-student overlaps
\begin{equation}\label{Sigma3}
    C(i,k)=\left(
    \begin{array}{ccc}
     Q_{ii} & Q_{ik}\\
     Q_{ik} & Q_{kk} \\
    \end{array}
    \right).
\end{equation}
One can solve the expectation values analytically for both activation functions with the integral terms $I_2$. For the error function, one finds for the generalization error \cite{saad1995line,goldt2019dynamics}
\begin{align}
\epsilon_g &=\frac{1}{\pi}\Biggl[\sum_{i,j} {c_i c_j \arcsin\hspace{-2pt}{\left(\frac{Q_{ij}}{\sqrt{1+Q_{ii}}\sqrt{1+Q_{jj}}\ }\right)}} \nonumber  \\[12pt]
&+\frac{1}{M} \sum_{n,m}{\arcsin\hspace{-2pt}{\left(\frac{T_{nm}}{\sqrt{1+T_{nn}}\sqrt{1+T_{mm}}\ }\right)}}\nonumber \\[15pt]
& -\frac{2}{\sqrt{M}}\sum_{i,n}{ c_i \arcsin\hspace{-2pt}{\left(\frac{R_{in}}{\sqrt{1+Q_{ii}}\sqrt{1+T_{nn}}\ }\right)}}\Biggr].
\label{final error2}\vspace{7pt}
\end{align}
and for a ReLU activation \cite{straat2019line}
\begin{align}
\epsilon_g =&\sum_{i,j} \frac{c_ic_j}{2} \biggl[\frac{Q_{ij}}{4}+ \nonumber \\ 
& \frac{1}{2\pi}\biggl[\sqrt{Q_{ii}Q_{jj}-Q_{ij}^2}+Q_{ij}\arcsin\left( \frac{Q_{ij}}{\sqrt{Q_{ii}Q_{jj}}}\right)\biggr] \nonumber \\
&-\sum_{i,n} \frac{c_i}{\sqrt{M}} \biggl[\frac{R_{in}}{4}+ \nonumber \\ 
& \frac{1}{2\pi}\biggl[\sqrt{Q_{ii}T_{nn}-R_{in}^2}+R_{in}\arcsin\left( \frac{R_{in}}{\sqrt{Q_{ii}T_{nn}}}\right)\biggr] \nonumber \\
&+\sum_{n,m} \frac{1}{2M} \biggl[\frac{T_{nm}}{4}+ \nonumber \\ 
& \frac{1}{2\pi}\biggl[\sqrt{T_{nn}T_{mm}-T_{nm}^2}+T_{nm}\arcsin\left( \frac{T_{nm}}{\sqrt{T_{nn}T_{mm}}}\right)\biggr]
\end{align}
For our setup $Q_{ii}=1, T_{nm}=\delta_{nm}$, and $M=1$, the expressions for the error activation and ReLU activation reduce to Eq. (\ref{final error reduced erf} )and Eq. (\ref{final error reduced relu}), respectively. Furthermore, we can also insert these definitions into Eq. (\ref{mean path}) in order to get the exact formula for the differential equations in both cases. 

\subsection{Relative variance estimation}
We find for the relative variance
\begin{align}
&\frac{\left<\frac{d\bm{c}}{d\alpha}^2\right>-\left<\frac{d\bm{c}}{d\alpha}\right>^2}{\left<\frac{d\bm{c}}{d\alpha}\right>^2} = \frac{\eta^2}{K}\frac{\left<\nabla\epsilon^2\right>-\nabla\epsilon_g^2}{\nabla\epsilon_g^2} \\
&= \frac{\eta^2}{K}\frac{\sum_{ij}^K \left<\left[\zeta-\sigma\right]^2\hspace{-2pt}g\hspace{-2pt}\left( x_i \right)g\hspace{-2pt}\left( x_j \right)\right>-\frac{4}{\pi^2}\left(\left(\tilde{Q}c-\tilde{R}\right)^2\right)_{\hspace{-3pt} ij}\hspace{-3pt}}{\frac{4}{\pi^2}\left(\left(\tilde{Q}c-\tilde{R}\right)^2\right)_{\hspace{-3pt} ij}}
\label{fluctuations2}
\end{align}
In general, we assume $c_i \sim \frac{1}{\sqrt{K}}$, that is at least given for the initialization. For the first part of the sum, we know that $ \left[\zeta-\sigma\right] \approx O\left(1\right)$ and assume that the student vectors are weakly correlated with joint variance of $\frac{1}{N}$. This gives
\begin{equation}
\sum_{ij}^K g\left( x_i \right)g\left( x_j \right) \approx O\left(\frac{K}{N}\right),
\end{equation}
as a sum over random variables with random signs. For the second part of the summation, we obtain
\begin{align}
\hat{c}_i=\sum_{j}^K \tilde{Q}_{ij}c_j &\approx \frac{1}{\sqrt{K}}\left(O\left(1\right)+\sum_{j\neq i}^K \tilde{Q}_{ij} \right) \nonumber \\
&\approx O\left(\frac{1}{\sqrt{K}}\right) + O\left(\frac{1}{\sqrt{N}}\right) ,
\end{align}
where the remaining sum for $j \neq i$ is a sum over random variables with zero mean and a variance of $ \frac{1}{N}$ . This leads to
\begin{align}
&\sum_{ij}^K \hat{c}_i \hat{c}_j \approx O\left(\frac{K}{N}\right),
\end{align}
as a sum over weakly correlated random variables. The last part leads to
\begin{align}
\sum_{ij}^K \left(\tilde{R}\tilde{R}^T\right)_{ij} \approx O\left(\frac{K}{N}\right).
\end{align}
Finally, one finds for the scaling of the relative variance
\begin{equation}
\frac{\left<\frac{d\bm{c}}{d\alpha}^2\right>-\left<\frac{d\bm{c}}{d\alpha}\right>^2}{\left<\frac{d\bm{c}}{d\alpha}\right>^2} \propto \frac{1}{K}.
\label{fluctuations3}
\end{equation}

\subsection{Details on the RFM with ReLU activation in the linearized regime}
Here, we want to derive the estimation of the following expectation value
\begin{align}
\left<\bm{R}^T \hat{\bm{Q}}^{-1}\bm{R}\right>=\frac{1}{N} \left< \text{tr}\left(\bm{Q} \hat{\bm{Q}}^{-1}\right)\right> .
\end{align},
with $\hat{\bm{Q}}^{-1}=\bm{A}^{-1} - \frac{\bm{A}^{-1} v v^T \bm{A}^{-1}}{2 \pi \left(1+\frac{1}{2\pi} v^T \bm{A}^{-1} v\right)}$. As mentioned in the main paper, this expectation value consists of two parts. The first part can be solved with similar steps as for the error activation function. In order to estimate $\lim_{K,N \rightarrow \infty} \left< \frac{1}{N}\text{tr}\left(\frac{\bm{Q} \bm{A}^{-1} v v^T \bm{A}^{-1}}{2 \pi \left(1+\frac{1}{2\pi} v^T \bm{A}^{-1} v\right)}\right)\right> $, we express $\bm{A}^{-1}$ with the help of the Woodbury matrix identity
\begin{align}
\bm{A}^{-1}&= \frac{4}{1-\frac{2}{\pi}} \left[\bm{1}_K-\frac{1}{1-\frac{2}{\pi}} \frac{\bm{J}^T}{\sqrt{N}} \left(\bm{1}_N+\frac{1}{1-\frac{2}{\pi}}\frac{\bm{J}\bm{J}^T}{N}\right)^{-1}\frac{\bm{J}}{\sqrt{N}}\right] \nonumber  \\
&= \frac{4}{1-\frac{2}{\pi}} \left[\bm{1}_K-\bm{C}\right]
\end{align}
with the student matrix $\bm{J} \in \mathbb{R}^{N\mathrm{x}K}$ and identity matrices in $K$-space $\bm{1}_K$ and $N$-space $\bm{1}_N$. Since $v v^T$ is a rank-1 matrix, the matrix $\bm{A}^{-1} v v^T \bm{A}^{-1}$ has still rank 1 and we can rewrite it
\begin{equation}
\bm{A}^{-1} \bm{v} \bm{v}^T \bm{A}^{-1} = \bm{u}\bm{u}^T
\end{equation}
with $u_i = \frac{4}{1-\frac{2}{\pi}} \left[1-\sum_l^K C_{il}\right]$. Therefore, we have $\bm{Q} \bm{A}^{-1} v v^T \bm{A}^{-1} =\bm{Q} \bm{u}\bm{u}^T$, which is a rank-1 matrix with one eigenvalue that is not zero and we estimate $\lambda=\bm{u}^T \bm{Q}\bm{u} = \sum_{ij} u_i Q_{ij}u_j \approx \sum_{i} u_i^2  = \mathcal{O}\left(K\right)  $ for the leading order of the sum. For the denominator, we obtain $\bm{v}^T \bm{A}^{-1} \bm{v} \approx \mathcal{O}\left(K\right)$ and the therefore the statement in the main text follows. 
\end{document}